\documentclass[sn-mathphys,Numbered]{sn-jnl}% Math and Physical Sciences Reference Style
\newcommand{\norm}[1]{\left\lVert #1 \right\rVert}
\usepackage{setspace}   % 用于控制行距
\pdfoutput=1
\usepackage[table]{xcolor}
\usepackage{booktabs}
\usepackage{graphicx}%
\usepackage{multirow}%
\usepackage{amsmath,amssymb,amsfonts}%
\usepackage{amsthm}%
\usepackage{mathrsfs}%
\usepackage[title]{appendix}%
\usepackage{textcomp}%
\usepackage{lmodern} % Better font coverage, fixes font shape warnings
\usepackage{manyfoot}%
\usepackage{booktabs}%
\usepackage{colortbl}
\usepackage{algorithm}%
\usepackage{algorithmicx}%
\usepackage{algpseudocode}%
\usepackage{listings}%
\usepackage{siunitx}
\usepackage{amssymb}
\usepackage{hyperref}
\usepackage{caption}
\usepackage{enumitem}
\usepackage{graphicx}%
\usepackage{multirow}%
\usepackage{amsmath,amssymb,amsfonts}%
\usepackage{amsthm}%
\usepackage{mathrsfs}%
\usepackage[title]{appendix}%
\usepackage{xcolor}%
\usepackage{textcomp}%
\usepackage{manyfoot}%
\usepackage{booktabs}%
\usepackage{algorithm}%
\usepackage{algorithmicx}%
\usepackage{algpseudocode}%
\usepackage{listings}%
\usepackage{afterpage}
\usepackage{float}
\usepackage{color}
\definecolor{forestgreen}{rgb}{0.33,0.61,0.34}

\begin{document}

\title{Toward generic control for soft robotic systems}

%%=============================================================%%
%% Prefix	-> \pfx{Dr}
%% GivenName	-> \fnm{Joergen W.}
%% Particle	-> \spfx{van der} -> surname prefix
%% FamilyName	-> \sur{Ploeg}
%% Suffix	-> \sfx{IV}
%% NatureName	-> \tanm{Poet Laureate} -> Title after name
%% Degrees	-> \dgr{MSc, PhD}
%% \author*[1,2]{\pfx{Dr} \fnm{Joergen W.} \spfx{van der} \sur{Ploeg} \sfx{IV} \tanm{Poet Laureate}
%%                 \dgr{MSc, PhD}}\email{iauthor@gmail.com}
%%=============================================================%%
\author[1,5]{\fnm{Yu} \sur{Sun}}

\equalcont{These authors contributed equally to this work.}

\author[2]{\fnm{Yaosheng} \sur{Deng}}

\equalcont{These authors contributed equally to this work.}

\author[3]{\fnm{Wenjie} \sur{Mei}}

\equalcont{These authors contributed equally to this work.}

\author*[1]{\fnm{Xiaogang} \sur{Xiong}}
\email{xiongxg@hit.edu.cn}

\author*[4]{\fnm{Yang} \sur{Bai}}
\email{yangbai@hiroshima-u.ac.jp}

\author[4]{\fnm{Masaki} \sur{Ogura}}

\author[1]{\fnm{Zeyu} \sur{Zhou}}

\author[2]{\fnm{Mir} \sur{Feroskhan}}

\author[5]{\fnm{Michael Yu} \sur{Wang}}

\author[6]{\fnm{Qiyang} \sur{Zuo}}
\author[7]{\fnm{Yao} \sur{Li}}

\author*[1]{\fnm{Yunjiang} \sur{Lou}}
\email{louyj@hit.edu.cn}

\affil[1]{\orgdiv{Shenzhen Key Lab of Advanced Motion Control Technology and Modern Automation Equipment},  \orgdiv{School of Intelligence Science and Engineering}, \orgdiv{College of Artificial Intelligence}, \orgname{Harbin Institute of Technology, Shenzhen}, \orgaddress{\street{HIT Campus, University Town of Shenzhen}, \city{Shenzhen}, \postcode{518055}, \state{Guangdong}, \country{China}}}

\affil[2]{\orgdiv{School of Mechanical and Aerospace Engineering}, \orgname{Nanyang Technological University}, \orgaddress{\street{50 Nanyang Avenue}, \city{Singapore}, \postcode{639798},  \country{Singapore}}}

\affil[3]{\orgdiv{School of Robotics and Automation, Suzhou Campus}, \orgname{Nanjing University}, \orgaddress{\street{1520 Taihu Road}, \city{Suzhou}, \postcode{215163}, \state{Jiangsu}, \country{China}}}

\affil[4]{\orgdiv{Graduate School of Advanced Science and Engineering}, \orgname{Hiroshima University}, \orgaddress{\street{1-4-1 Kagamiyama}, \city{Higashi-Hiroshima}, \postcode{739-8527}, \state{Hiroshima}, \country{Japan}}}

\affil[5]{\orgdiv{School of Advanced Engineering}, \orgname{Great Bay University}, \orgaddress{\street{No. 16 Daxue Road, Songshan Lake High-tech Industrial Development Zone}, \city{Dongguan}, \postcode{523000}, \state{Guangdong}, \country{China}}}

\affil[6]{\orgdiv{Center for Precision Engineering}, \orgname{Shenzhen Institutes of Advanced Technology, Chinese Academy of Sciences}, \orgaddress{\street{1068 Xueyuan Avenue, University Town of Shenzhen}, \city{Shenzhen}, \postcode{518055}, \state{Guangdong}, \country{China}}}

\affil[7]{\orgdiv{School of Robotics and Advanced Manufacture}, \orgdiv{College of Artificial Intelligence}, \orgname{Harbin Institute of Technology, Shenzhen}, \orgaddress{\street{HIT Campus, University Town of Shenzhen}, \city{Shenzhen}, \postcode{518055}, \state{Guangdong}, \country{China}}}

%%==================================%%
%% sample for unstructured abstract %%
%%==================================%%

\abstract{Soft robotics has advanced rapidly, yet its control methods remain fragmented: different morphologies and actuation schemes still require task-specific controllers, hindering theoretical integration and large-scale deployment. A generic control framework is therefore essential, and a key obstacle lies in the persistent use of rigid-body control logic, which relies on precise models and strict low-level execution. Such a paradigm is effective for rigid robots but fails for soft robots, where the ability to tolerate and exploit approximate action representations, i.e., control compliance, is the basis of robustness and adaptability rather than a disturbance to be eliminated.
Control should thus shift from suppressing compliance to explicitly exploiting it. Human motor control exemplifies this principle: instead of computing exact dynamics or issuing detailed muscle-level commands, it expresses intention through high-level movement tendencies, while reflexes and biomechanical mechanisms autonomously resolve local details. This architecture enables robustness, flexibility, and cross-task generalization.
Motivated by this insight, we propose a generic soft-robot control framework grounded in control compliance and validate it across robots with diverse morphologies and actuation mechanisms. The results demonstrate stable, safe, and cross-platform transferable behavior, indicating that embracing control compliance, rather than resisting it, may provide a widely applicable foundation for unified soft-robot control.}

\maketitle

\section{Introduction}

With the rapid advancement of flexible materials and bio-inspired structures, soft robots have expanded the functional boundaries of traditional robotic systems. Their exceptional deformability, safe physical interaction, and ability to navigate complex environments make them ideal for minimally invasive surgery~\cite{abdelaziz2024fiberbots,york2021microrobotic,min2024stiffness}, post-disaster rescue~\cite{bai2025swarm,montagut2024cyborg}, and traversal of unstructured terrains. Yet the very compliance that endows these capabilities simultaneously exposes the limits of existing control paradigms. Soft-robot dynamics are high-dimensional, nonlinear, and time-varying, rendering analytical modeling and prediction exceedingly difficult~\cite{Goury2018soft,huang2020dynamic,Thuruthel2019soft,deng2025safety}. As a result, existing controllers are often customized for particular robot morphologies and actuation mechanisms, leaving the field without a unified control framework.

\begin{figure}[H]
    \centering
    \includegraphics[width=0.8\linewidth]{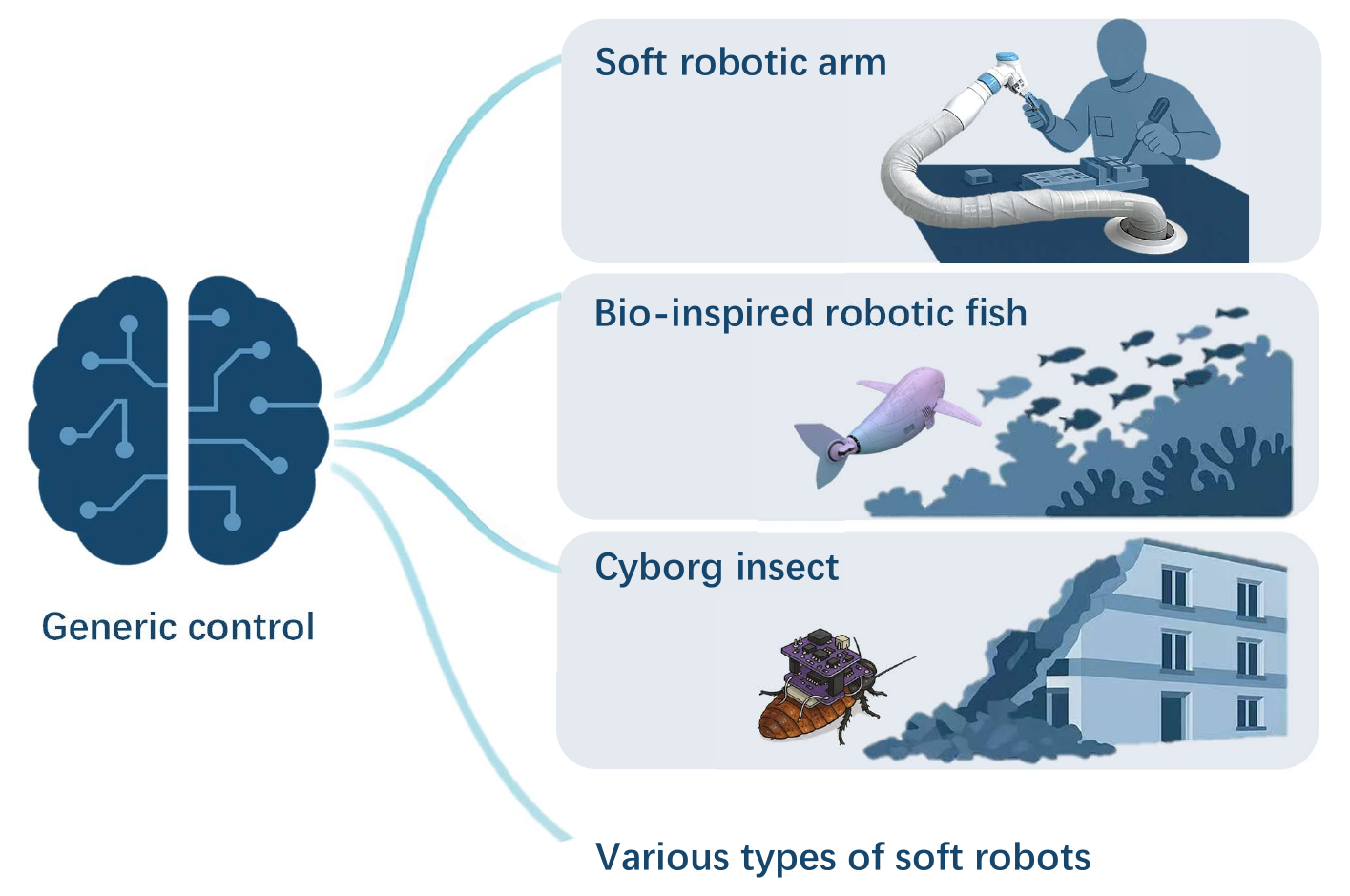}
    \caption{\textbf{Generic control framework for soft robots.}  This study presents a generic and safety-guaranteed control framework that enables unified control of diverse soft robotic systems. The unified framework is applicable to soft robotic arms, bioinspired fish robots, cyborg insects, as well as other soft robots operating across various environments, demonstrating morphology-independent control.
}
    \label{fig:fig1}
\end{figure}
% \newpage

The pursuit of such a framework holds deep engineering and scientific significance. Engineering-wise, today’s controllers are tightly coupled to specific material distributions and actuation layouts, preventing reuse and hindering scalable deployment. Scientifically, treating each robot as an isolated system obscures the fundamental commonalities underlying bending, coordinated deformation, and environmental adaptation. A structure-independent, robust control principle is therefore essential not only for practical generalization but also for uncovering universal coordination laws in soft systems.

Existing approaches fall into three major categories. Embodied control paradigms encode behavioral logic directly into physical structure or materials, for example through asymmetric geometries, magnetic responsiveness, or material intelligence~\cite{Rus2015Nature,Wehner2016Nature,Dong2022SciAdv}. But because these behaviors are hard-wired at design time, adaptability is inherently limited. Data-driven methods approximate complex dynamics via neural or learning-based models, offering adaptability at the cost of massive data requirements, heavy computation, and weak safety guarantees~\cite{Thuruthel2018SoRoSurvey,Thuruthel2019TRO,Shih2020SciRob}. Theoretical control methods provide formal stability but depend critically on accurate models and incur computational overhead~\cite{Renda2018TRO,Goury2018TRO,Huang2020NatComm}. Despite steady progress, these paradigms remain fundamentally fragmented: none achieve generalizable and robust control across varying morphologies and environments~\cite{Rus2015Nature,Thuruthel2018SoRoSurvey,Shih2020SciRob}.

A central cause of this fragmentation could be philosophical rather than technical. For decades, soft-robot control has inherited the rigid-body paradigm, built on precise models and strict execution. These methods construct detailed geometric, actuation, and dynamical models, then derive specialized control laws accordingly. While effective for rigid robots, such a reliance on precision becomes untenable when applied to soft bodies whose morphology and mechanics are deeply entangled.
More importantly, soft robots do not require, and cannot benefit from strict low-level precision. What they truly need is approximate, high-level expression of action. For example, in soft robotic fish, the controller only needs to specify an overall left-right oscillation trend, rather than controlling the exact phase of the tail at each instant; the fluid–structure interaction naturally resolves the detailed motion and forms a stable propulsion pattern. Since local behaviors in soft robots are regulated automatically by the interaction between system dynamics and the environment, strict low-level commands are neither necessary nor feasible.

For this reason, we introduce the concept of control compliance, which captures the essential requirement of soft-robot control: Control compliance refers to the ability of a control system to tolerate approximate action representations, generating high-level tendencies or intentions rather than precise low-level commands, while allowing local details to be resolved autonomously by system dynamics and environmental interactions.
Control compliance fundamentally contradicts the precision assumptions of rigid-body control: the former treats moderate inaccuracy as a resource to be exploited, while the latter treats any inaccuracy as an error to be eliminated. Consequently, imposing rigid-body control logic on soft systems not only proves ineffective but suppresses the natural self-regulation and adaptability that give soft robots their advantages.

Soft-robot control should therefore shift from suppressing control compliance to exploiting it. Achieving this requires control principles that do not rely on accurate modeling or fine-grained actuation, yet remain stable and generalizable under highly complex dynamics. The human neuromuscular system offers a compelling example: humans do not compute precise global dynamics or issue explicit muscle-level commands; instead, intention is expressed through high-level movement tendencies, while biomechanical properties and reflexes automatically regulate local details. This strategy of “controlling tendencies rather than controlling details” provides robustness, flexibility, and cross-task generalization.
Inspired by this principle, we propose a generic control framework comprising three modules: a learning module that generates movement tendencies, a safety module that regulates these tendencies under control-compliance constraints, and a planning module that maps the tendencies into executable inputs.

To evaluate this framework, we consider three representative classes of systems: (i) continuum robots such as soft arms, modeled with constant-curvature or Cosserat-rod dynamics~\cite{Webster2010IJRR,Renda2018TRO}; (ii) environment-coupled systems such as soft fish or underwater manipulators, where motion arises from strong interactions with fluids or granular media~\cite{Katzschmann2018SciRob,Sinatra2019SciRob,Naclerio2021SciRob,Li2023NatComm}; and (iii) biohybrid systems such as tissue-engineered or cyborg robots, where uncertainty arises from biological variability~\cite{bai2025swarm,Ricotti2017SciRob,lin2025cyborg,Park2016Science,Kakei2022npjFlexElec}. These categories span the dominant dynamical regimes of soft robotics, enabling comprehensive cross-platform validation. Experimental results demonstrate that the proposed framework achieves stable, predictable, and safe control across diverse morphologies and environments, confirming its robustness and transferability under uncertainty. More importantly, the results reveal a broader implication: soft-robot control can and should leverage compliance, as compliance itself constitutes one of the most robust and transferable control resources available.

\clearpage

\section{Results}
This section details the proposed generic control framework to achieve reliable and transferable control across soft robotic systems. The controller design is first introduced, followed by experimental validation on three distinct platforms: a tendon-driven soft arm, a soft-bodied robotic fish, and a cyborg cockroach.
\subsection{Controller design}

This study proposes a generic control framework for soft robots, addressing the long-standing lack of unified, safe, and generalizable control approaches in this field. The control framework is inspired by how humans control their bodies. Human motor control operates effectively in dynamic and uncertain environments without relying on precise dynamic models~\cite{Wolpert1995Science,Todorov2002NatNeuro}. Instead, it is driven by three key mechanisms: coarse internal representations of body–environment interactions~\cite{Wolpert1995Science,Kording2004Nature}, rapid sampling-based action selection grounded in perception and experience~\cite{Scott2004NatRevNeuro}, and reflexive safety regulation triggered near boundaries of safe motion~\cite{Cullen2019NatRevNeuro,Jamali2016NatComm}. Humans update their intended motion in a step-by-step manner rather than generating full trajectories in advance, and they maintain stability by steering away from unsafe regions rather than strictly converging toward preset targets. These observations highlight a crucial insight: in complex soft systems, greater model precision does not necessarily yield better control performance; fast, adaptive, and coarse-grained strategies could sometimes be more effective.

\begin{figure}[H]
    \centering
    \includegraphics[width=1\linewidth]{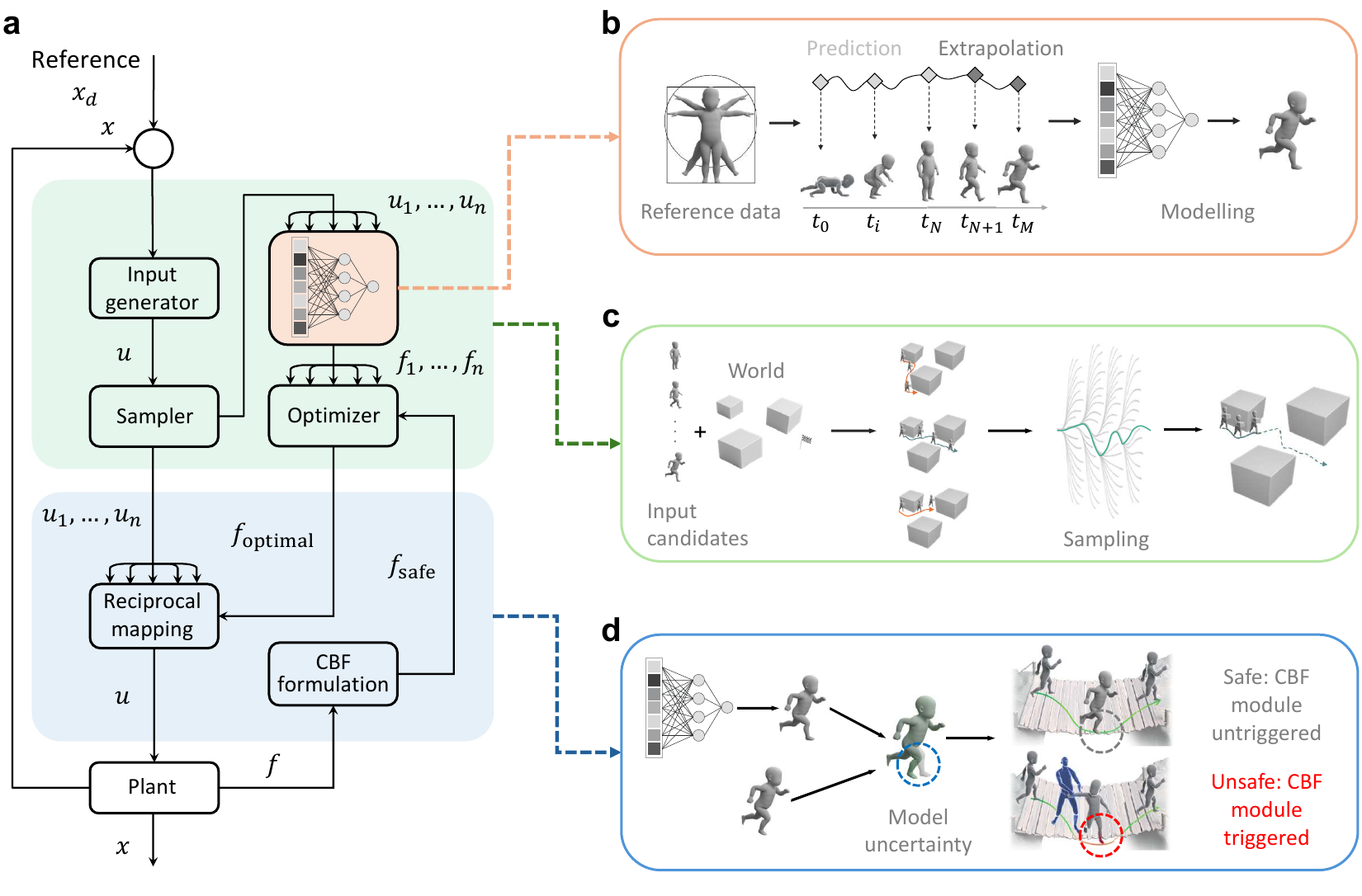}
    \caption{
  \textbf{Overview of the proposed control framework inspired by human motor intelligence}. \textbf{a} Overall architecture. The control framework consists of three core modules: a learning-based model, a sampling-based planner, and an adaptive safety filter. The learning-based model employs neural ordinary differential equation (Neural ODE) to approximate the intrinsic robot–environment dynamics; the sampling-based planner generates candidate control sequences via sampling-based model predictive control (SBMPC) formulation and predicts corresponding outputs; and the adaptive safety filter, grounded on control barrier function (CBF), selects the optimal predicted motion within the provably safe region and maps it back to real-time control signals through a reciprocal mapping. This structure mirrors the human motor control process (coarse modeling, rapid planning, and reflex-like safety regulation) to maintain both flexibility and stability under uncertainty.
 \textbf{b} Learning-based model. The Neural ODE captures dynamics of the soft robot and provides bounded prediction errors for motion forecasting, similar to how the human brain forms and refines internal representations of body–environment interactions through perception and experience.
 \textbf{c} Sampling-based planner. The SBMPC-based formulation explores and optimizes candidate motion sequences using importance sampling, achieving efficient planning of soft robotic motion, much as the human nervous system performs rapid sample-based motion planning in complicated environments.
 \textbf{d} Adaptive safety filter. The adaptive CBF defines a provably safe region for the soft robot and converts the selected optimal prediction into stable, executable control inputs, ensuring robust and safe behavior under external perturbations. This mechanism resembles how the human nervous system triggers reflexive corrections when motion approaches safety limits, much like a parent catching a falling child.}
    \label{fig:2}
\end{figure}

Motivated by this principle, we propose a generic soft robot control framework that unifies learning-based modeling, sampling-based planning, and adaptive safety regulation (Fig.~\ref{fig:2}). The learning module provides a coarse yet adaptable representation of robot–environment interactions; the sampling-based planner selects feasible actions in a stepwise, human-like manner; and the adaptive safety mechanism implements reflex-like protection when approaching unsafe regions. Working together, these components preserve the foundational intent of soft robotics and enable robust, fast, and generalizable control performance in complex and uncertain environments, achieving levels of safety, adaptability, and environmental interaction that rigid-body methods inherently struggle to deliver.
\newpage

\subsubsection{Learning-based model}

{Our learning-based model was based on the observation of how humans learn to walk (Fig.~\ref{fig:2}b). In early development, a child first learns to crawl, gradually stabilizing its center of mass and coordinating limb movements. Through crawling, the body acquires a sense of balance and learns to maintain stability even under continuous motion. When transitioning from crawling to kneeling and finally to standing, this prior experience enables the body to stabilize a higher and dynamically changing center of mass, eventually achieving stable walking.}

{This process mirrors the function of our Neural ODE network in the learning-based model. During training, the Neural ODE fits the boundary conditions and observed reference data of the soft robotic system, learning a physically consistent representation of its dynamics. At each point along the prediction time series $t_0, \ldots, t_N$, the model not only predicts the potential outcome of an action but also takes an exploratory step into previously unseen dynamic regions. Beyond $t_N$, it can further extrapolate outside the training distribution to nearby regions, showing that what the Neural ODE ``remembers" is not mere data samples but the underlying physical consistency embedded in the learned dynamics.}

\subsubsection{Sampling-based planner}

{Through learning-based modeling, humans learn to walk toward a goal in an open environment. However, when the surroundings become crowded, the brain begins to imagine several short trajectories a few steps ahead, evaluating each imagined path and selecting the one that appears most reliable at that moment. After taking the first step, the brain perceives how the surroundings have changed, imagines a few new paths, evaluates them, and moves again—continuously repeating this perception and decision cycle to make its way through complex spaces (Fig.~\ref{fig:2}c).}

{Similarly, the sampler (Fig.~\ref{fig:2}a) first explores a large number of candidate motion sequences through importance sampling, and each sequence is propagated through the Neural ODE model to obtain the corresponding motion prediction. The predicted trajectories are then evaluated by incorporating environmental information and weighted exponentially according to their cost functions. The control sequence with the optimal weighted cost is selected as the next-step command. Only the first action of this sequence is executed before the system updates its perception, generates new samples, and repeats the process, thereby enabling rapid and adaptive decision-making for soft robots in complex environments.}

\subsubsection{Adaptive safety filter}

{While the sampling-based controller enables fast and efficient decision-making, potential safety risks remain due to the coarse learning of the model and environmental variations. To address these uncertainties, we introduce a safety filter that rapidly evaluates the sampled control inputs and corrects unsafe ones before execution.}

{This design is inspired by the way parents accompany a child learning to walk. When the child moves safely along the path, the parents simply observe from behind without interference. However, as the child approaches the edge of a dangerous area, the parents immediately intervene, gently guiding the child back onto a safe path (Fig.~\ref{fig:2}d). Similarly, the adaptive CBF  defines a provably safe region for the soft robot and converts the selected optimal prediction into stable and executable control inputs. The CBF remains passive when observing motion $f$ from the soft system inside the safety region. However, it becomes active instantly when the predicted motion $f$ approaches the safety boundary. Like parents quickly correcting a child's movement, CBF formulation provides a safe motion $f_{\text{safe}}$ to the optimizer, and then calculating an optimal $f_{\text{optimal}}$ to select the corresponding candidate control signal $u$ by reciprocal mapping, which ensures robust and stable behavior even under external perturbations.
This mechanism resembles how the human nervous system triggers reflexive corrections
when motion approaches safety limits, much like a parent catching a falling child.}

\subsubsection{Reciprocal mapping based cross-module coordination}

It is important to note that these three modules cannot be simply connected in sequence, because none of them can directly generate executable control inputs. The learning-based dynamics model is a black box, making it difficult to construct control laws on top of it. The SBMPC module can explore feasible trajectories, but searching for control inputs through it is computationally expensive and unsuitable for real-time use. More critically, traditional CBF methods require explicit and accurate system dynamics to provide formal safety guarantees, a requirement fundamentally incompatible with a black-box model.

The clue to addressing this limitation also comes from human motor control. When performing a movement, humans do not compute individual muscle torques or joint angles. Instead, they first form a high-level movement tendency: a coarse motion sketch such as ``extend the arm forward" or ``shift the body to the right to maintain balance." This tendency is not the final motor command but an intermediate representation that lower-level systems refine and execute. Human motion, therefore, emerges through a gradual tendency-to-execution process rather than through direct low-level input design.

Our control architecture adopts this idea. The learning module first generates surrogate actions that express movement tendencies. All interactions between the planning and safety modules then take place on this surrogate-action level rather than the control-input level: the safety module filters the surrogate actions using adaptive CBF constraints to obtain a safe subset, and the sampling-based planner retrieves or generates executable control inputs from this safe subset. In this structure, the learning module provides the feasible dynamic space (movement-tendency generation), the safety module defines its admissible region (boundary shaping), and the planning module maps tendencies to executable inputs (final execution). The three modules communicate through the surrogate action, and a reciprocal-mapping block establishes a bidirectional correspondence between surrogate actions and real control inputs, enabling coarse movement tendencies to be reliably converted into concrete motor commands. This mapping mechanism allows the three modules to coordinate through a shared intermediate representation, achieving generalized, safe, and transferable control across diverse soft robotic platforms.

\subsection{Deployment on diverse soft robotic systems}
To evaluate the generality and applicability of our proposed controller, we deploy it across three structurally and functionally distinct soft robotic platforms: a tendon-driven soft arm, a soft-bodied robotic fish, and a cyborg cockroach. These platforms differ significantly in morphology, dynamics, actuation methods, and operational environments, ranging from structured planar settings to unstructured aquatic and biological domains. The successful deployment across all three systems confirms that the controller accommodates diverse embodiments and maintains reliable, safety-aware performance across varying tasks and constraints.

\subsubsection{Tendon-driven soft arm}
\begin{figure}[!htbp]
    \centering
    \includegraphics[width=1\linewidth]{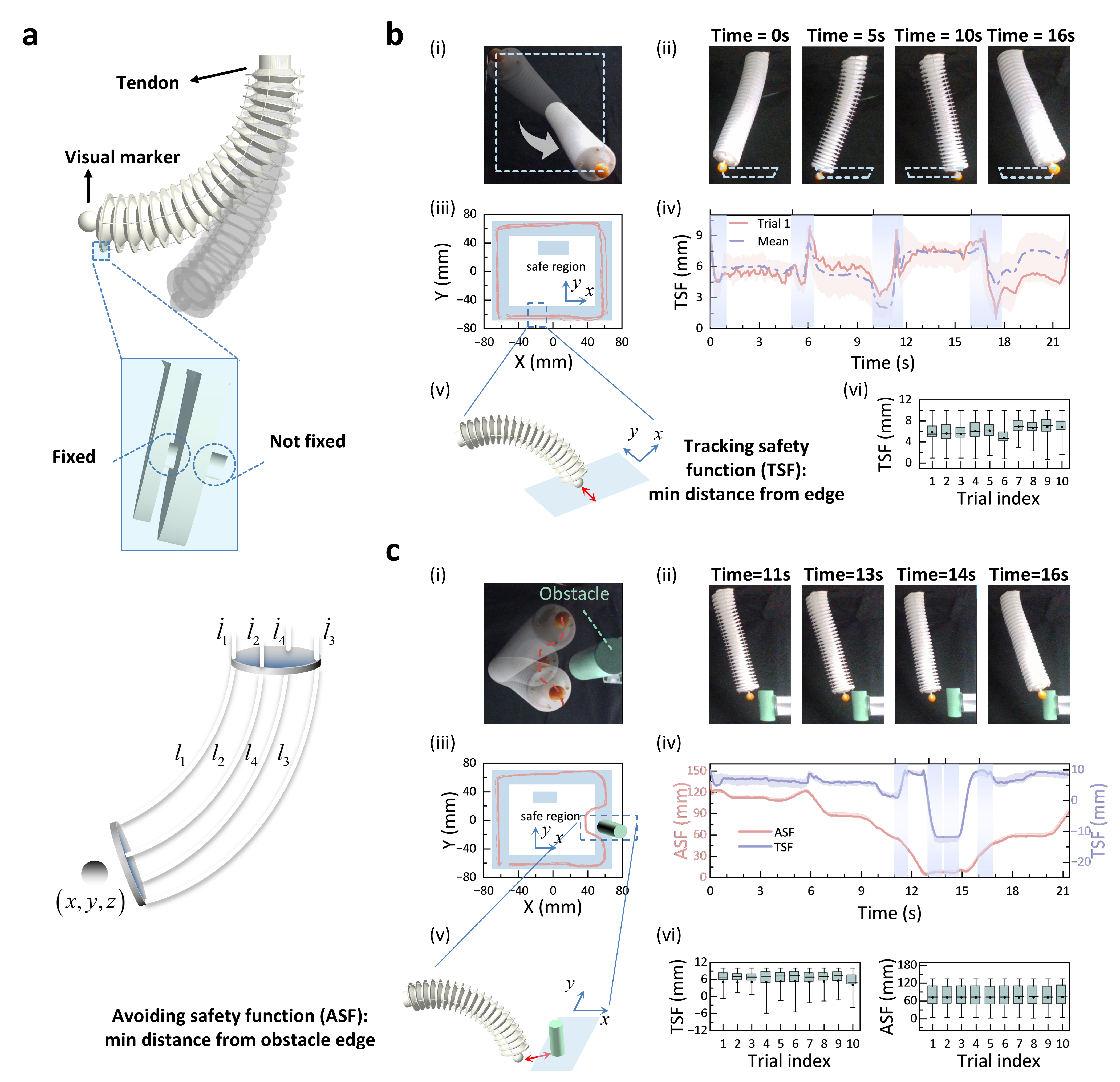}
    \caption{\textbf{Experimental validation on the soft robotic arm. The experiments include trajectory tracking and obstacle avoidance of the arm's end-effector.}
    \textbf{a} Structure and actuation of the tendon-driven soft arm.
    \textbf{b} Square trajectory tracking with safety constraints. Top and side views are shown in (i) and (ii). The end-effector follows a square path within the defined safety margin (blue band) (iii). (iv) shows the temporal evolution of the tracking safety function (TSF), defined as the minimum distance to the boundary (see (v)). It indicates safe operation when within 0-10. As shown in (iv), TSF remains stable over time, with slight dips at corners but no violations. Boxplots in (vi) show consistent TSF distributions across ten trials, with all trajectories remaining within the safe region.
    \textbf{c} Obstacle-avoidance performance. Top and side views in (i) and (ii) show the arm adapting its motion to avoid a green obstacle, with corresponding trajectories in (iii). The temporal evolution of TSF and the avoidance safety function (ASF) is shown in (iv), where ASF is defined as the minimum distance to the obstacle (see (v)). It remains positive throughout, ensuring safe obstacle avoidance, when TSF is relaxed to satisfy ASF constraints. Boxplots in (vi) show stable TSF and ASF distributions, confirming the controller’s consistency and reliability.}

    \label{fig:fig3}
\end{figure}

We first evaluated the proposed generic controller on a tendon-driven soft arm to assess its performance on soft manipulators with elastic compliance and actuation redundancy. The arm is commanded to follow a square trajectory in a planar workspace while avoiding a fixed obstacle. 

In the trajectory tracking task, as shown in Fig.~\ref{fig:fig3}b, the end-effector consistently follows the square path across 10 repeated trials. The tracking safety function (TSF), defined as the minimum distance to the region boundary, remains strictly positive throughout the trials, with a minimum value of~$0.70$. These results indicate robust tracking performance under safety constraints.

In the obstacle avoidance experiment, as shown in Fig.~\ref{fig:fig3}c, the controller successfully guides the arm around a fixed cylindrical obstacle. The avoiding safety function (ASF), defined as the minimum distance between the arm and the obstacle, remains above~$2.71$ across all trials. During the obstacle avoidance phase, ASF is prioritized to ensure collision-free execution, resulting in a temporary decrease in TSF. Once the obstacle is cleared, both safety functions return to stable and safe values.

These experiments validate that the controller achieves accurate, consistent, and safety-critical motion in tendon-driven soft arms, confirming its generalizability across soft robotic platforms with high degrees of compliance.
\clearpage

\subsubsection{Soft-bodied robotic fish}
To further evaluate the generality of our proposed controller, we deployed it on a soft-bodied robotic fish performing autonomous obstacle avoidance in a confined aquatic environment. The fish features a compliant and multi-segment tail actuated by a single servo motor, enabling undulatory locomotion. Two fixed spherical obstacles are placed in the water tank.

\begin{figure}[p]
    \centering
    \includegraphics[width=1\linewidth]{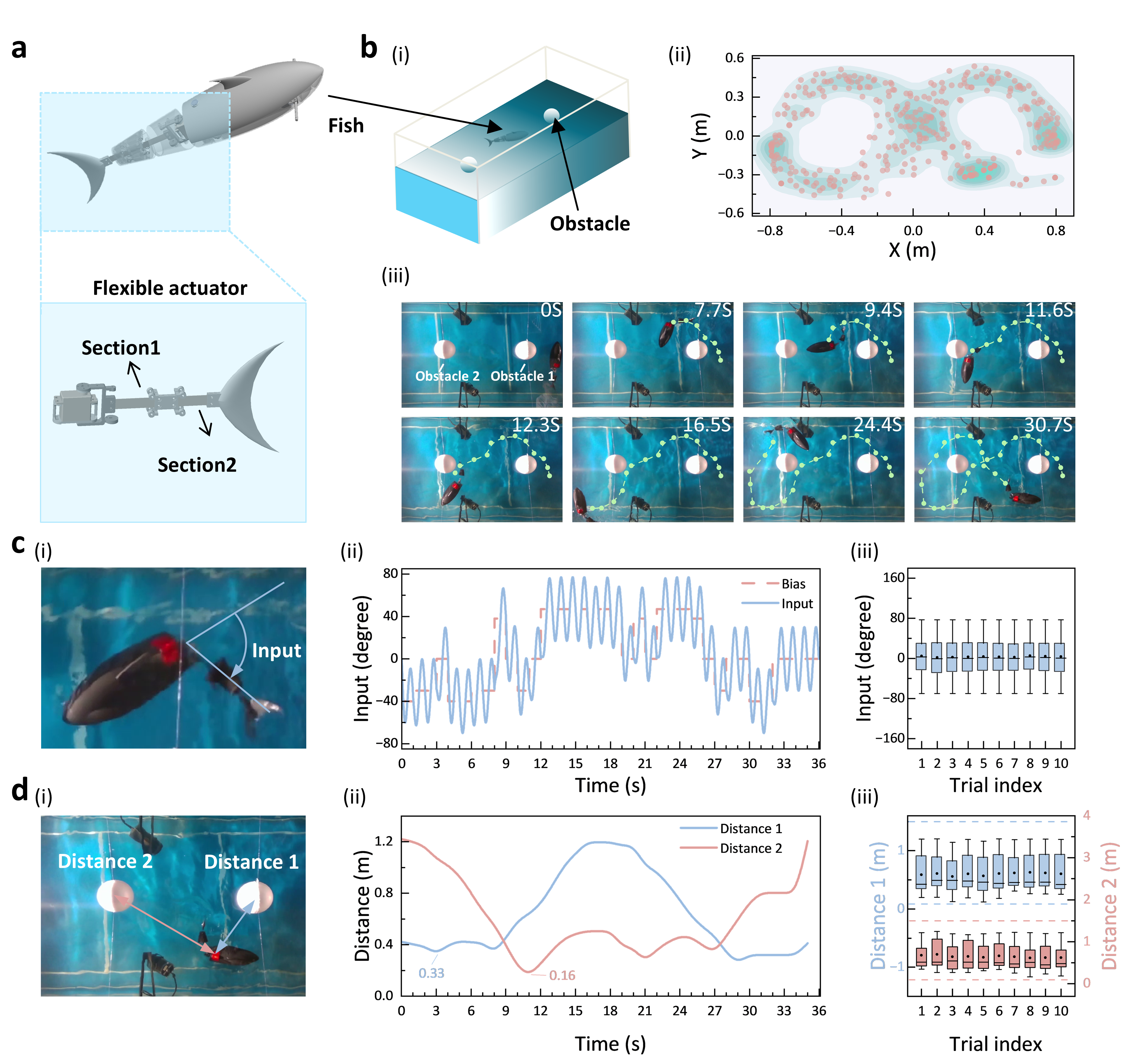}
    \caption{\textbf{Experimental validation on a soft-bodied robotic fish navigating a confined aquatic environment.}
    \textbf{a} Structure of the soft-bodied robotic fish propelled by a multi-segment flexible tail.
    \textbf{b} Navigation in an obstacle-filled environment. The setup with two obstacles is shown in (i). The robot followed a figure-eight trajectory over ten trials, as seen in (ii), successfully avoiding collisions. Snapshots in (iii) illustrated smooth swimming of a representative experiment trial 1.
    \textbf{c} Turning control through input modulation. The control input is defined as the angle between the body and first tail segment (i). A sinusoidal signal governs tail motion, and turning is achieved by modulating its bias; the input trajectory for trial 1 is shown in (ii). Boxplots in (iii) confirm that the required tail oscillation amplitudes remain below the designed safety limit.
    \textbf{d} Evaluation of obstacle distances. Distances to obstacle 1 (blue) and 2 (red) are defined in (i). In trial 1 (ii), both remain above the safety threshold with minima of 0.33 m and 0.16 m. Boxplots in (iii) confirm safety over ten trials.
    }
    \label{fig:fig4}
\end{figure}

As shown in Fig.~\ref{fig:fig4}b, the robotic fish successfully completed ten consecutive trials under the proposed controller. A trajectory density plot reveals a consistent figure-eight pattern around the obstacles, indicating reliable repeatability and adaptive maneuvering capability. Time-stamped snapshots from a representative trial (Fig.~\ref{fig:fig4}b-iii) further illustrate smooth and collision-free navigation.

The ability to turn (i.e., change heading direction) is achieved by modulating the bias of the input signal, defined as the angle between the fish body and the first tail segment (Fig.~\ref{fig:fig4}c). Varying this bias enables left and right turns based on the planned motion. Experimental results confirm that stable directional control is achieved with minimal input modulation.

To assess safety, we tracked the Euclidean distance between the fish and each obstacle throughout the motion. In a representative trial (Fig.~\ref{fig:fig4}d), the minimum distances to the two obstacles were $0.33\ \mathrm{m}$ and $0.16\ \mathrm{m}$, respectively, both comfortably exceeding the predefined safety threshold of 0.10 m. Boxplots summarizing the minimum distances across all ten trials (Fig.~\ref{fig:fig4}d-iii) further confirm that distances consistently remained well within the safe range, underscoring robust and reliable obstacle avoidance in a cluttered aquatic environment.

These results demonstrate that the proposed generic controller enables autonomous, safe, and highly repeatable operation across morphologically distinct soft robotic platforms, including the soft robotic arm and the fish-like robot in constrained environments.

\clearpage

\subsubsection{Cyborg cockroach}

To evaluate the applicability of our generic control framework to soft bio-hybrid platforms, we conducted straight-line locomotion experiments using a cyborg cockroach. The system consists of a live Madagascar hissing cockroach equipped with a custom electronic backpack that delivers directional electrical stimulation to induce turning. The goal is to guide the cockroach to a predefined target while keeping its trajectory within a narrow safety corridor.

\begin{figure}[p]
    \centering
    \includegraphics[width=0.9\linewidth]{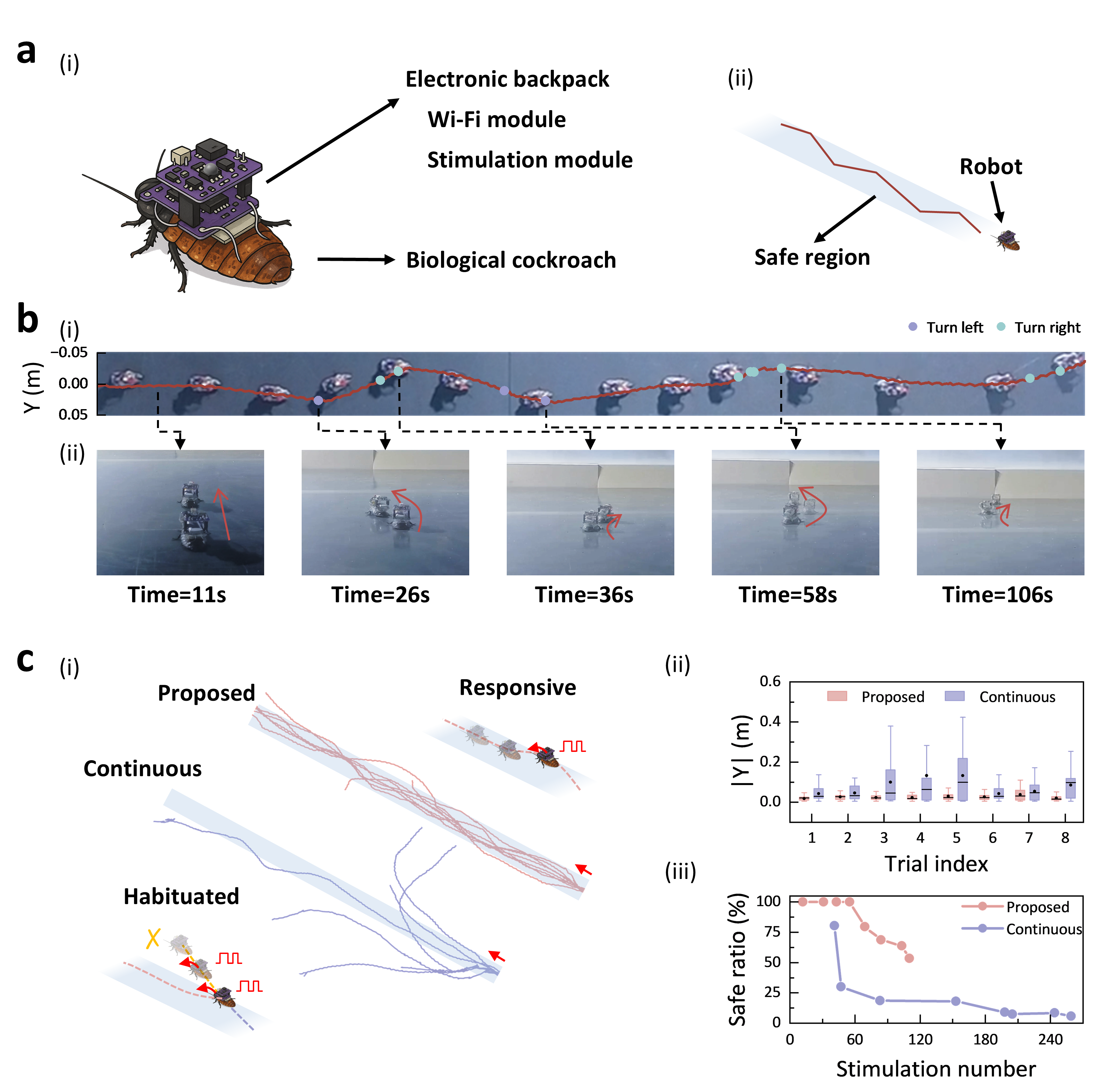}
    \caption{\textbf{Experimental validation on a cyborg cockroach performing straight-line locomotion}. 
    \textbf{a} Structure and task setup of the cyborg cockroach.The composition of the cyborg cockroach is shown in (i). The cockroach follows a straight-line path within the defined safety margin (blue band) (ii).
    \textbf{b} Straight-line locomotion with safety constraints. Top and rear views are shown in (i) and (ii), with stimulation events labeled for left turns (purple) and right turns (cyan). The cockroach remains within the safety region throughout.
    \textbf{c} Comparison with continuous stimulation strategy. (i) shows trajectories under both algorithms. With the proposed event-triggered strategy, most trajectories stay within the safety region (blue band). Under continuous stimulation, trajectories often exceed this region due to habituation. Boxplots in (ii) show the $|Y|$ distribution under both algorithms. The proposed method yields a stable distribution with lower variance and a smaller mean. As the number of stimulations increases, habituation emerges in both cases, leading to a decline in safety ratio, as shown in (iii).
    }  
    \label{fig:fig5}
\end{figure}

As shown in Fig.~\ref{fig:fig5}a, the platform includes a lightweight wireless stimulation module mounted on the cockroach (Fig.~\ref{fig:fig5}a-i). The task environment defines a straight path with a lateral safety margin of $\pm 0.05\ \mathrm{m}$ (Fig.~\ref{fig:fig5}a-ii). The task is considered successful if the trajectory remains entirely within this corridor after the cockroach arrived at its predefined target.
Fig.~\ref{fig:fig5}b presents a representative trial under our control strategy. The top-view trajectory (orange-red) shows that the cockroach stays within the safety region throughout. Stimulation events for left and right turns are shown as purple and cyan dots, triggered only when the trajectory approaches the corridor boundaries. Rear-view snapshots illustrate key moments, including straight walking, initial turning, and subsequent corrections. These results confirm that the controller achieves reliable navigation with less interventions.

We further compare our proposed method with a continuous stimulation strategy in which electrical stimuli are persistently applied whenever the heading deviates from the desired direction. As shown in Fig.~\ref{fig:fig5}c-i, under our controller, most trajectories remain within the designated safety region even as the number of stimulations increases. In contrast, the continuous stimulation controller frequently produces trajectories that violate the safety boundaries.
The boxplot in Fig.~\ref{fig:fig5}c-ii shows the distribution of $|Y|$ deviations under both methods. Our approach keeps lateral deviations consistently bounded across eight trials, whereas the continuous strategy results in significantly larger deviations.
Fig.~\ref{fig:fig5}c-iii summarizes the relationship between the number of stimulations and the safety ratio, which is defined as the proportion of the trajectory that remains within the safety corridor. Under continuous stimulation, the safety ratio decreases rapidly as the number of stimulations increases. This effect is primarily due to the cockroach's habituation to repeated stimuli. In comparison, the proposed method maintains a high safety ratio with fewer interventions even when the cumulative stimulation count exceeds 120.

\clearpage

These results demonstrate that the proposed generic controller is effective and robust when deployed on soft bio-hybrid platforms. By triggering interventions only when necessary for safety, the method achieves reliable task performance with fewer stimulations, reducing the risk of overstimulation and slowing down habituation to stimulation in living agents.

\clearpage
\section{Discussion}
To further validate the generality and practicality of the proposed framework, we conducted comparative experiments covering long-term operation, physical constraints, and dynamic environments. The results demonstrate that the framework exhibits remarkable adaptability, stability, and real-time efficiency when facing challenges such as model degradation and input limitations, highlighting its advantages in soft robotic control.
\begin{figure}[!htbp]
    \centering
    \includegraphics[width=0.9\linewidth]{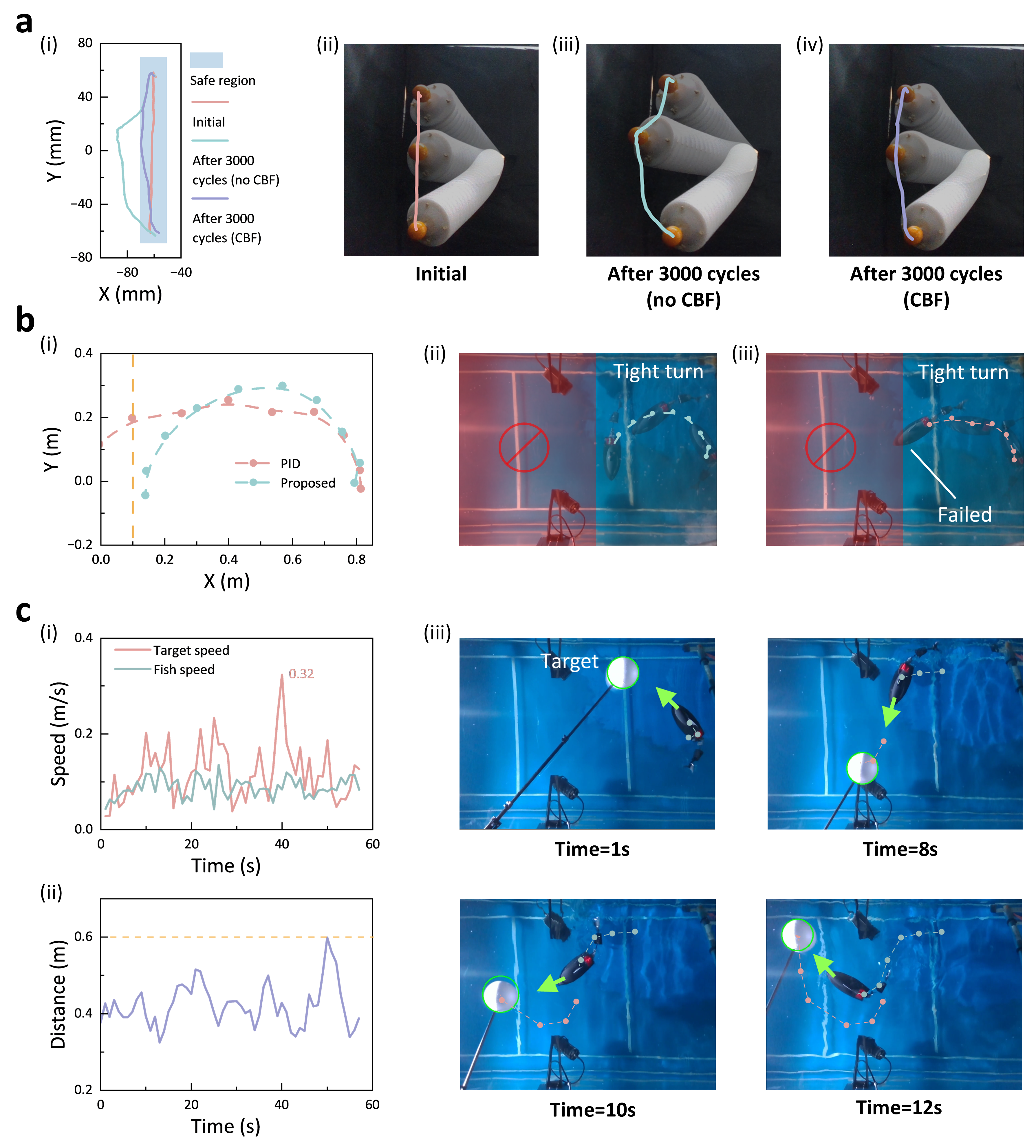}
    \caption{\textbf{Experimental validation of the framework's robustness, safety, and real-time performance}. \textbf{a} Robustness against model degradation in long-term operation. The purpose was to test our algorithm's ability to maintain performance despite material fatigue. Initially, the controller tracks the desired trajectory accurately within the safe region (i, red line; ii). After 3,000 cycles, the controller without the adaptive CBF fails, causing the manipulator to violate the predefined safe region (i, cyan line; iii). In contrast, our framework with the adaptive CBF enabled compensates for the model degradation, successfully maintaining safe and accurate performance (i, purple line; iv). \textbf{b} Safe navigation under physical constraints. This experiment aimed to demonstrate the algorithm's capacity to handle actuator input limits during challenging maneuvers. A robotic fish was tasked with executing a tight turn in a confined space, comparing our method against a standard PID controller. The PID controller fails the maneuver due to input saturation (i, red dashed line; iii). Our framework successfully navigates the tight turn by generating constraint-compliant control actions, keeping the fish on its intended path (i, cyan dashed line; ii). \textbf{c} Real-time control in a dynamic environment. This experiment validated the algorithm's efficiency in real-time tracking of a speed-varying target using a robotic fish. The results demonstrate successful real-time tracking, as the fish's speed closely matched the target's speed, which reached a maximum of $0.32~$m/s (i, iii). The controller consistently maintained a close tracking distance throughout the 60-second trial (ii).}
    \label{fig:fig6}
\end{figure}
\subsection{Long-term performance maintenance}

A significant challenge in deploying soft robots is their performance degradation over extended periods. This occurs because factors like material fatigue cause the robot's dynamics to change, making initial models inaccurate. We demonstrate this problem and our solution in a long-term experiment with a soft manipulator over {three thousands} operational cycles (Fig.~\ref{fig:fig6}a). Initially, the controller accurately tracks the desired trajectory (Fig.~\ref{fig:fig6}a-ii, red line). After three thousands cycles, the performance of the controller without our adaptive CBF module degrades substantially, causing the manipulator to violate its predefined safe region (Fig.~\ref{fig:fig6}a-i,~\ref{fig:fig6}a-iii, cyan line). In contrast, by enabling the adaptive CBF, our framework compensates for the model degradation and keeps the manipulator within the safe region, successfully maintaining its initial performance (Fig.~\ref{fig:fig6}a-i,~\ref{fig:fig6}a-iv, purple line).

The key to this sustained performance is our method's different approach to handling model uncertainty. Instead of pursuing an ever-more-perfect, but ultimately brittle, dynamic model, our framework is designed to operate effectively with an imperfect model by directly accounting for uncertainty and prediction errors. The framework has two main components. First, the Neural ODE module learns the robot's core dynamics from data to create a predictive model. Second, the CBF module provides a safety layer. It continuously measures the error between the model's predictions and the robot's actual behavior, and uses this information to update the safety constraints in real-time.
This mechanism allows the controller to compensate for the gradual degradation of the learned model, enables reliable, long-duration autonomy for soft robots in unstructured environments.
\subsection{Safe control under physical constraints
}

To operate safely in the real world, soft robots must respect physical constraints, such as actuator input limits, to prevent hardware damage. This is directly related to the need for smooth control inputs, as sudden changes can cause destructive internal stresses. Integrating these constraints is a major challenge for traditional controllers. We demonstrate this in a navigation task where a robotic fish must execute a tight turn in a confined space (Fig.~\ref{fig:fig6}b). As shown in the experiment, a PID controller with a simple input saturation limit is unable to complete the maneuver, causing the fish to fail the turn (Fig.~\ref{fig:fig6}b-i,~\ref{fig:fig6}b-iii, red line). In contrast, our framework successfully navigates the tight turn, keeping the fish on the desired path while respecting the same input limits (Fig.~\ref{fig:fig6}b-i,~\ref{fig:fig6}b-ii, cyan line).

This successful performance is achieved through the synergy between the optimizer and the safety module in our framework. A traditional PID controller performs poorly when its input is limited due to issues like integral windup. Our method works differently: the SBMPC module generates a diverse set of possible control actions. The adaptive CBF module then acts like a filter, ensuring that the final selected control input is not only effective for the task but also smooth and compliant with all physical constraints.

\subsection{Real-time control in dynamic environments}
Another feature of our framework is its computational efficiency, which enables real-time control of soft robots in dynamic environments. We demonstrated this in a dynamic target-tracking experiment, where a robotic fish was tasked to follow a moving target with fluctuating speeds up to 0.32 m/s (Fig.~\ref{fig:fig6}c). The results show that the fish's speed closely matched the target's speed in real-time (Fig.~\ref{fig:fig6}c-i), and the controller successfully maintained a close tracking distance throughout the experiment (Fig.~\ref{fig:fig6}c-ii). The experimental snapshots further visualize this successful tracking performance (Fig.~\ref{fig:fig6}c-iii).

This real-time capability is due to the framework's sampling-based control architecture. Unlike conventional MPC, which must solve a computationally expensive optimization problem at every time step, our SBMPC module explores the solution space efficiently by sampling a limited number of candidate trajectories. This reduces the per-step computation time, making real-time implementation feasible. More importantly, this structure is inherently parallelizable: each sampled trajectory can be evaluated independently, allowing computation to be distributed across multiple cores. This means that the framework can scale gracefully: future scenarios that require increased sampling density or more complex evaluations can be handled without a prohibitive increase in latency.

\clearpage

\bibliography{reference}  % No \bibliographystyle

\clearpage

\section{Method}\label{sec1}
\subsection{Control framework design}

\subsubsection{Learning-based model}

To construct this model, we follow a three-stage learning procedure. First, we collect a set of input–output trajectories from the physical system under diverse actuation patterns and environmental conditions, forming a representative dataset that reflects the robot's deformation responses. Second, we embed a neural network inside the ODE formulation to parameterize the unknown dynamics as
\begin{equation}\label{sys-nctrl}
\dot{x}(t) = f(x(t), u(t)) + \theta(t),
\end{equation}
where state $x\in\mathcal{X}\subset \mathbb R^n$, input $u\in \mathbb R^m$. $f\colon \mathbb R^n\times \mathbb R^m\to\mathbb R^n$ is a Lipschitz function. $\theta\in\Theta\subset \mathbb R^n$ represents the external uncertainty. Here, $f$ represents the desired internal dynamics of the system, which correspond to the underlying physical model without considering external disturbances
$\theta(t)$.

After collecting the training data, we trained the Neural ODE parameters by minimizing the trajectory-level discrepancy between predicted and measured motion. Then, the trained Neural ODE approximates the underlying system dynamics and predicts the robot’s motion response under arbitrary control inputs.

Since the Neural ODE learns only an approximate representation of the system, modeling uncertainty remains inevitable. To quantify this uncertainty and enable safe integration into control, we characterize an explicit upper bound on the modeling error. Let $f(x(t), u(t))$ denote the Neural ODE prediction and $f_{\text{des}}(t)$ a reference trajectory representing the desired system behavior. The prediction error $\varepsilon(t)=f(x(t),u(t)) - f_{\text{des}}(t)$ is bounded by a calibrated constant $\delta_{\max}\ge0$, such that $|\delta(t)|\le\delta_{\max}$ holds within the range of states and inputs observed in the data~\cite{mei2024controlsynth}. Hence, the learned Neural ODE is equipped with a certified upper bound on its prediction error, quantifying the worst-case discrepancy between the model and the true dynamics.

{With the error bound established, the model becomes sufficiently trustworthy for control despite not being an exact physical description. This reflects the intended coarse-modeling philosophy of our Neural ODE design, where the goal is not to replicate the full analytical physics of the soft robot. Instead, it captures the essential nonlinear trends and physical regularities embedded in the observed data, providing a coarse but coherent description of robot–environment interactions.}

\subsubsection{Sampling-based planner}

Building upon the learned Neural ODE dynamics, the SBMPC module makes real-time decisions by exploring multiple candidate control sequences to the Neural ODE and evaluating their predicted outcomes. The formal description of these stages is presented below.

First, at time step $t_k$, the sampler operates within a finite prediction horizon of length $H$. A set of candidate control sequences $U =\{u_1,\ldots, u_n\}$ is drawn from a parameterized probability distribution and propagated through the Neural ODE to generate the corresponding predicted motion sequences $F = \{f_1,\ldots,f_n\}$, enabling the controller to rapidly explore feasible short-horizon motions without relying on explicit analytical models.
Then, the SBMPC evaluates each predicted sequence $F$ by optimizing a composite cost that reflects both motion accuracy and safety:

\begin{equation}\label{OCP}
J(F, U) = \sum_{k=1}^H \ell(x_k, u_k) + \lambda(x_k),
\end{equation}
where $x_k$ and $u_k$ denote the state and candidate control at time step $k$, $\ell(x_k,u_k)$ measures the tracking error with respect to the desired motion, and $\lambda(x_k)$ balances optimization performance and penalties.

{Among all candidates, the prediction $f_i$ that minimizes $J(F,U)$ in~\eqref{OCP} is selected as the provisional optimal prediction $f_{\text{optimal}}$, i.e., $f_{\text{optimal}} = \arg\min_{f_i\in F}J(F,U)$. This optimal prediction serves as a data-driven estimate of the desired next-step motion, representing the planner's best prediction under the learning-based model.
This provisional prediction $f_{\text{optimal}}$ is then provide to the adaptive safety filter, which evaluates and refines it to ensure consistency with the system's safety constraints.}

\subsubsection{Adaptive safety filter}

{The sampling-based planner provides optimal prediction motion $f_{\text{optimal}}$, which  contains approximation errors from model uncertainty. To ensure safety under these uncertainties, we introduce an adaptive CBF module as a reflex-like safety filter. This module evaluates whether the provided optimal prediction $f_{\text{optimal}}$ is within an admissible safety region and intervenes only when necessary.}

{To formalize this safety region, we design $\mathcal{C}\subset\mathbb{R}^n$ to dynamically adjust its boundary according to model uncertainty. The safety condition is defined by a CBF $h(x)$, which characterizes the admissible set of system states satisfying $h(x)\ge 0$. Based on this formulation, the adaptive safety region is constructed as 
\begin{equation}\label{kcbf}
\begin{aligned}
\mathcal{C} 
=\{ x_k\mid \Phi\!+\Pi\geq 0\},
\end{aligned}
\end{equation}
}
{where $\Pi = f_{\text{optimal}}^\top\dfrac{\partial h}{\partial x_k} \! \norm{\dfrac{\partial h}{\partial x_k}}\bar\varepsilon$ is a control-related term that reflects the effect of the prediction $f_{\text{optimal}}$ on safety set~\eqref{kcbf}, $\Phi =  \left(\hat \theta_k  \!+\!\dfrac{\partial h}{\partial \hat\theta_k}\right)^\top \dfrac{\partial h}{\partial x_k}+ h
$ is an adaptive term that compensates for model uncertainty, and the adaptive term $\hat \theta$ is updated by $\dot{\hat \theta} = -\partial h/\partial x$. The CBF $h(x)$ in $\Pi$ and $\Phi$ defines the admissible state set that satisfies $h(x)\ge0$.}

With the safety region $\mathcal{C}$ defined, the adaptive CBF evaluates whether the planner's provisional prediction is admissible. If $f_{\text{optimal}}\in \mathcal{C}$ the motion is considered safe and is passed directly to the execution layer. 
However, if $f_{\text{optimal}}\notin \mathcal{C}$, {the safety filter intervenes in a reflex-like manner. In this case, the CBF formulation first extracts all predicted motions that within the safety region, forming the set $\mathcal{S} = \mathcal{C} \cap \{f_1,\ldots,f_n\}$. Then, among these admissible candidates, the safety filter selects a prediction $f_{\text{safe}}$ that is closest to the provisional optimal one $f_{\text{optimal}}$, i.e.,
\begin{equation}
f_{\text{safe}} = \arg\min_{f_i \in \mathcal{S}} \|f_i - f_{\text{optimal}}\|. \end{equation}
}
{This selected $f_{\text{safe}}$ is then used in the subsequent reciprocal mapping process to determine the corresponding executable control $u_{\text{safe}}$, which ensures the executed command remains consistent with both the learned dynamics and the safety constraints.}

\subsubsection{Reciprocal mapping based cross-module coordination}

{Once the adaptive safety filter provides the verified safety prediction $f_{\text{safe}}$, the reciprocal mapping searches within the candidate control sequence $\{u_1,\ldots,u_n\}$ provided by the sampler to find the corresponding control input $u_{\text{safe}}$.}

{This mapping explicitly encodes the correspondence between the predicted motion sequence $\{f_1,\ldots,f_n\}$ and the candidate control sequence $\{u_1,\ldots,u_n\}$ established by the sampler and Neural ODE. It serves as a bridge that allows the system to retrieve the control sequence associated with a given safe prediction. Since $f_{\text{safe}}$ is selected by the adaptive safety filter from the predicted motion sequence $\{f_1,\ldots,f_n\}$, it follows that $f_{\text{safe}} \in \{f_1,\ldots,f_n\}$ and there exists a one-to-one correspondence mapping between the two sets $\{f_1,\ldots,f_n\}$ and $\{u_1,\ldots,u_n\}$. Therefore, the system can retrieve the corresponding control sequence $u_{\text{safe}}$ associated with the safety-verified motion $f_{\text{safe}}$ through the reciprocal mapping.} {Formally, we define the reciprocal mapping as
\begin{equation}
u_{\text{safe}} = \mathcal{M}(f_{\text{safe}}), \quad f_{\text{safe}} \in \{f_1,\ldots,f_n\},\quad u_{\text{safe}} \in \{u_1,\ldots,u_n\},
\end{equation}
where $\mathcal{M}(\cdot)$ denotes a bidirectional mapping function that constructed directly from the one-to-one pairs $(u_i, f_i)$ generated by the sampler and Neural ODE forward propagation.} 

{This reciprocal mapping preserves the internal consistency between the learned dynamics and control inputs, ensuring that safety-verified prediction can be transferred into executable control input without additional computation.}

\subsection{Robotic systems}

\subsubsection{Tendon-driven soft arm}

The soft arm shown in Fig.~\ref{fig:fig3} was fabricated as a single monolithic structure via 3D printing with thermoplastic polyurethane (TPU), which confers high axial compliance while preserving sufficient torsional stiffness for precise control. The arm measures 300~mm in length and 48~mm in diameter, thereby optimizing the balance between structural integrity and flexibility. Four tendons are symmetrically routed through the arm and anchored at the distal tip, enabling omnidirectional bending in three-dimensional space. Each tendon is independently actuated by a NEMA 17 stepper motor via a low-backlash capstan-drive mechanism. The end-effector bears a visual fiducial marker that is tracked in real time by an external Intel RealSense D435i depth camera, yielding 3D pose feedback for performance evaluation and closed-loop control.

\subsubsection{Soft-bodied robotic fish}

The soft robotic fish depicted in Fig.~\ref{fig:fig4} incorporates an undulatory propulsion system actuated by a flexible caudal tail. The robot measures approximately 450~mm $\times$ 90~mm $\times$ 94~mm and weighs 0.88kg. The rigid head houses all onboard electronics, including a wireless communication module (APC220), an inertial measurement unit (JY901B), a 32-bit microcontroller (STM32F407ZET6), and a 14.8V lithium-polymer battery.

The caudal tail comprises two serially connected passive joints fabricated from spring steel, which facilitate smooth and continuous bending under actuation. A single waterproof servo motor (GDW-BLS896) imparts periodic oscillations at the tail base, generating a propagating traveling wave that yields forward thrust. The body and tail were molded from soft silicone-based elastomers via a multi-part casting process, ensuring watertightness and compliant deformation under hydrodynamic loads. 

This tail-driven propulsion system endows the fish robot with exceptional maneuverability and seamless navigation within confined aquatic environments, such as narrow corridors and bottleneck spaces.

\subsubsection{Cyborg cockroach}
\paragraph{Experimental species and ethics.}

The adult \textit{Gromphadorhina portentosa} shown in Fig.~\ref{fig:fig5} were housed at approximately 25\,\textdegree C and 60\% relative humidity, with food and water provided \textit{ad libitum}. All experimental procedures were approved by the Institutional Animal Care and Use Committee (IACUC) of Harbin Institute of Technology, Shenzhen (approval ID: IACUC-2020026). Prior to surgery, animals were briefly anesthetized with CO$_2$ for approximately one minute. Platinum microelectrodes (bare diameter: 76.2\,\textmu m; insulated: 139.7\,\textmu m) were implanted bilaterally into the optic lobes at a depth of 2\,mm and secured using biocompatible adhesive.

\paragraph{Electronic backpack and stimulation system.}

A lightweight backpack was mounted on the insect's thorax, shown in Fig.~\ref{fig:fig5}a, integrating power supply, wireless control, and stimulation modules. The system includes a Wi-Fi-enabled microcontroller, a DC--DC voltage regulator, and a constant-current stimulator (Intan V063) communicating via SPI. The stimulator supports both unidirectional (0--2550\,\textmu A) and bidirectional (0--5100\,\textmu A) modes, with resolution of 1\,\textmu A below 255\,\textmu A and 10\,\textmu A above. Stimulation protocols included forward-drive pulses (50\,Hz, 40--160\,\textmu A) and turning pulses (50\,Hz, 80--110\,\textmu A), applied to either the abdomen or optic lobe depending on the task. The total backpack weight, including battery, is approximately 7.1\,g.

\clearpage

\section{Acknowledgements}

This paper is partially supported by the National Natural Science Foundation of China, grant No. 62573159,  Guangdong Provincial Key Laboratory of Intelligent Morphing Mechanisms and Adaptive Robotics under Grant  2023B1212010005 and in part by Shenzhen Science and Technology Program under Grant KJZD20240903100802004  and GXWD 20231130153844002.
\section{Author contributions} 
Y. B. and X. X. conceived and designed the research. Y. B., Y. D., and Y. S. developed the control framework. Y. S., Z. Z., Q. Z., and Y. L. built the experimental platform and robots. Y. S. and Z. Z. conducted the experiments and collected the raw data. Y. S., Z. Z., Y. B., Y. D., X. X., and W. M. conducted the data analysis. Y. B., Y. D., Y. S., X. X., W. M., Y. J. L., M. Y. W., M. F., and M. O. contributed to the preparation of the manuscript. X. X., Y. B., and Y. J. L. supervised the research.
\section{Competing interests} The authors declare no competing interests.

\end{document}